\def\year{2020}\relax
\documentclass[letterpaper]{article} 
\usepackage{aaai20}  
\usepackage{times}  
\usepackage{helvet} 
\usepackage{courier}  
\usepackage[hyphens]{url}  
\usepackage{graphicx} 
\usepackage{graphicx}  
\usepackage[colorinlistoftodos]{todonotes}
\usepackage{amssymb}
\usepackage{float}
\usepackage{amsmath}
\usepackage{ulem}
\usepackage[ruled,vlined]{algorithm2e}
\newcommand{\norm}[1]{\left\lVert#1\right\rVert}
\DeclareMathOperator*{\argmax}{argmax}

\title{Wavelets to the Rescue: Improving Sample Quality of Latent Variable Deep Generative Models}
\author{\textbf{Prashnna Kumar Gyawali}\textsuperscript{1},
\textbf{Rudra Saha}\textsuperscript{2}, \textbf{Linwei Wang}\textsuperscript{1}, \textbf{VSR Veeravasarapu}\textsuperscript{2}, \textbf{Maneesh Singh}\textsuperscript{2}\\
\textsuperscript{\rm 1} Rochester Institute of Technology, Rochester, NY 14623, USA\\
\textsuperscript{\rm 2} Verisk AI, Verisk Analytics\\
{pkg2182@rit.edu, rudra.fnu@verisk.com, linwei.wang@rit.edu, vsr.veera@gmail.com, maneesh.singh@verisk.com}
}
 
\begin{document}

\maketitle

\begin{abstract}
Variational Autoencoders (VAE) are probabilistic deep generative models underpinned by elegant theory, stable training processes, and meaningful manifold representations.
However,  they  produce blurry images due to a lack of explicit  emphasis over high-frequency textural details of the images, and the difficulty to directly model the complex joint probability distribution over the high-dimensional image space.
In this work, we approach these two challenges with a novel wavelet space VAE 
that uses the decoder to model the images in the wavelet coefficient space. 
This enables the VAE to emphasize over high-frequency components within an image obtained via wavelet decomposition. 
Additionally, by decomposing the complex function of generating high-dimensional images into inverse wavelet transformation and generation of wavelet coefficients, the latter becomes simpler to model by the VAE. 
We empirically validate that deep generative models operating in the wavelet space can generate images of higher quality than the image (RGB) space counterparts. 
Quantitatively, on benchmark natural image datasets, we achieve consistently better FID scores than VAE based architectures and competitive FID scores with a variety of GAN models for the same architectural and experimental setup.
Furthermore, the proposed wavelet-based generative model retains desirable attributes like disentangled and informative latent representation without losing the quality in the generated samples. 
\end{abstract}

\section{Introduction}
In recent years, deep generative models such as Variational Autoencoders (VAEs)  \cite{kingma2013auto,rezende2014stochastic}, Generative Adversarial Networks (GANs) \cite{goodfellow2014generative}, and their variants have been studied extensively. 
GANs have primarily succeeded in synthesizing high-resolution photographic images but are still facing challenges in training stability and sampling diversity \cite{karras2019style}. VAEs, on the other hand, are theoretically elegant, easier to train, and have meaningful manifold representations and yet they have not been able to attain a similar performance, with their generated samples lacking finer details \cite{huang2018introvae,dai2019diagnosing}.

The inability of VAEs to produce high-quality realistic images can be attributed to two challenges. 
Firstly, VAE objective often includes pixel-wise mean squared errors for auto-encoding, resulting in overly smooth images that lack textural details. 
While the concept of \textit{texture} lacks a formal definition, it generally refers to small-scale patterns of variations of visual stimulus which has been analysed classically using spatial-frequency analysis, for example, wavelet analysis,
in signal processing literature \cite{datta2008image}. 
Furthermore, a correlation between 
the  amount  of  frequency  content  in  natural  images and the 
widely-used image quality metric of Fr\'echet Inception Distance (FID) \cite{heusel2017gans} is also found in this work. This drives our hypothesis that utilizing spatial-frequency analysis of images for building VAE models may help mitigate the above issue.

Secondly, VAEs aim to capture the joint probability distribution function, $f(\mathbf{x})$, of the image $\mathbf{x}$. 
For high dimensional image domain, since such $f(\mathbf{x})$ is not known or is too complex to specify, the approximation of $f(\mathbf{x})$ via generative models may be difficult. The difficulties in modeling such a complex function, however, can be relaxed utilizing the idea of transform-domain models.
These models restructure the image using a linear  invertible transform, leaving coefficients whose structure is simpler to model.
From this perspective, we investigate if building the generative models for the coefficients of such well-recognized transform-domain models can help in better modeling of $f(\mathbf{x})$ and thus increasing the sample quality.

To approach these two challenges, we rely on well-known wavelet transform which possesses attractive properties for natural image modeling \cite{mallat1989theory,mallat2012group}. Wavelet decomposition consists of high-frequency details $\mathbf{w}_{hf}$ along with a low-frequency approximation of images $\mathbf{u}_{lf}$. This can be utilized to build generative models for high frequency along with low-frequency components.
Further, since inverse wavelet transforms $\mathcal{I}_{wt}(\mathbf{w}_{hf}, \mathbf{u}_{lf})$ are loss-less,
generative models for wavelet coefficients 
 $f_{wt}(\mathbf{w}_{hf}, \mathbf{u}_{hf})$ reduces the complexity of modeling the complex image based model. Image generation, then, is a two step process: sampling wavelet coefficients $\mathbf{w}_{hf}, \mathbf{u}_{hf} \sim f_{wt}(\mathbf{w}_{hf}, \mathbf{u}_{hf})$, followed by (inversely) transforming coefficients to image-space, $x = \mathcal{I}_{wt}(\mathbf{w}_{hf}, \mathbf{u}_{lf})$.

In this work, we propose a wavelet space generative model based on variational inference and utilize the inverse wavelet transform to generate image samples from the generated wavelet coefficients. We compare the proposed method with standard VAE architectures on three benchmark natural image datasets: CIFAR-10 \cite{krizhevsky2009learning}, CelebA \cite{liu2015deep} and Flickr-Face dataset \cite{karras2019style}. Across all these datasets, we achieve improved sample quality of generated samples, which we compare both qualitatively and quantitatively. 
We also compare the use of wavelet-based generative model against image-based GAN for similar architectural setup and achieve competitive performance.
Further, we also investigate the attributes of standard VAE like disentangled and informative latent representation. Qualitatively, the proposed wavelet-based generative model exhibits the disentanglement without losing sample quality. Additionally, compared to VAE, the proposed wavelet-based generative model obtains superior mutual information measure.

Overall, our contributions are as follows:
\begin{itemize}
\item We propose a wavelet-based VAE for improving the sample quality of generated images compared to standard VAE. 
\item We demonstrate increased sample quality, compared to standard VAE, both qualitatively and quantitatively on three benchmark datasets: CIFAR-10, CelebA, and Flickr-Face. 
\item We achieve competitive results when replacing the image-space generator, within a variety of GAN models, with the wavelet-space generator
for the same architectural and experimental setup.
\item We evaluate the properties of disentanglement and information embedded in latent spaces of the proposed wavelet-based generative models.
\end{itemize}


\section{Related Work}
Recently, there has been an effort in understanding and improving the image generation capability of VAE \cite{dai2019diagnosing,heljakka2018pioneer,huang2018introvae}. In 
the recently proposed two-stage VAE \cite{dai2019diagnosing}, 
the mismatch of latent representation with the sampling prior 
was identified as the main limiting factor 
for the generation capability of VAE. 
As a remedy, they proposed the training of a second-stage VAE to model the latent representation learned by the first-stage VAE. The work in \cite{heljakka2018pioneer} modified the training strategy of the VAE to grow the network architecture progressively, an approach previously 
shown to be effective for GAN \cite{karras2017progressive}. 
Other works such as \cite{huang2018introvae} 
modified the architecture and ELBO loss objective 
to achieve high-resolution image synthesis. Unlike these works, we neither introduce a second-stage training nor change the training or architecture style of VAE. 

Wavelet transforms for natural images are well-studied in signal processing literature \cite{mallat1989theory,mallat2012group}, 
used widely in a vareity of problems including texture modeling, 
denoising, and detection in natural images \cite{crouse1998wavelet,romberg2001bayesian}. The recent use of wavelets in deep generative models has been mostly around GAN \cite{liu2019attribute,angles2018generative}. 
In \cite{liu2019attribute}, a wavelet-based hybrid GAN 
was proposed, which includes an image-space generator and a wavelet-space discriminator to emphasize facial attributes resulting due to aging.
In \cite{angles2018generative}, a wavelet-based generator was proposed where the generative model was computed by inverting a fixed scattering embedding of the training data bypassing the requirement of a discriminator for optimization. 
Unlike these works, 
we focus on improving VAE with wavelets which, to the best of our knowledge, is the first VAE that explores the generative modeling in the wavelet space.
Furthermore, we define a
probabilistic model for the wavelets via our VAE decoder (\textit{i.e.}, generator) and then perform the inversion.



\begin{figure}[tb]
\begin{center}
  \includegraphics[width=0.47\textwidth]{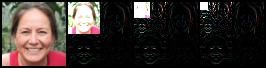}
\end{center}
\caption{\small{Left: The original image obtained from \cite{karras2019style}. Right: Decomposition of the image at several scales using Wavelet Transform.}} 
\label{fig:wavletDecomposition}
\end{figure}

\section{Preliminaries}
\subsection{
Wavelet Transform}
\label{sec:waveletTransform}
Wavelet transform is  well studied for statistical signal and image processing. Its primary properties include: \textit{locality} (represents image content local in space and frequency), \textit{multi-resolution} (represents image at varying set of scales),  and \textit{texture detection} (acts as local edge detectors) \cite{romberg2001bayesian}. A wavelet $\mathbf{\psi} \in \mathbf{L}^{2}(\mathrm{R}^d)$ is dilated with a scale sequence \{$a^{j}\}_{j\in \mathrm{Z}}$ for $a > 1$, to construct a wavelet transform, where the scale $a$ is typically 2 for image processing \cite{mallat2012group}. The 2D discrete wavelet transform represents an image $\mathbf{x}$ in terms of scaling functions (also known as approximation coefficients) \{$\mathbf{\psi}^{LL}$\} and a set of shifted and dilated wavelet functions \{$\mathbf{\psi}_{b}$\}, where $b \in \mathcal{B} := $\{LH, HL, HH\} denotes the subbands of wavelet decompositions. This is represented as:
\begin{equation}
    \mathbf{x} = \sum_{k\in\mathrm{Z}^2}u_{j_0, k}\cdot\psi^{LL}_{j_0, k} + \sum_{b\in\mathcal{B}}\sum_{j\geq j_0}\sum_{k\in\mathrm{Z}^2}w^{b}_{j, k}\cdot\psi^{b}_{j, k}
\end{equation}
where $u$ and $w$ are approximation and wavelet coefficients respectively. In Fig. ~\ref{fig:wavletDecomposition}, the wavelet decomposition of an image at three different scales is shown. The approximation coefficients are placed in the left top patch and rest of the patches in each level are the wavelet coefficients representing the structure and textural information of the facial image. 

\subsection{
Variational Autoencoder}
Variational Autoencoders \cite{kingma2013auto,rezende2014stochastic} are latent variable models that describe a stochastic process by which the data being modeled is assumed to be generated. It assumes that data $\mathbf{x}$ 
is generated by latent variable $\mathbf{z}$ where 
the likelihood $p_{\theta}(\mathbf{x}|\mathbf{z})$ 
is parameterized by a neural network. 
To allow inference of both $\mathbf{z}$ and $\theta$, 
a recognition model $q_{\phi}(\mathbf{z}|\mathbf{x})$, also parameterized by a neural network, is introduced to approximate the intractable true posterior $p_{\theta}(\mathbf{z}|\mathbf{x})$. 
This gives rise to an encoding-decoding structure,  
where the probabilistic encoder 
describes $q_{\phi}(\mathbf{z}|\mathbf{x})$ and 
the probabilistic decoder describes $p_{\theta}(\mathbf{x}|\mathbf{z})$. 
The objective of VAE
maximizes the variational evidence lower bound (ELBO) 
of the marginal data likelihood with respect to network parameters $\theta$ and $\phi$:
\begin{equation}
  \label{eq:VAEObj}
  \log{p}(\mathbf{x}) \geq \mathcal{L}  = \mathop{\mathbb{E}_{q_{\phi}(\mathbf{z}|\mathbf{x})}}[\log{p_{\theta}(\mathbf{x}|\mathbf{z})}] - KL(q_{\phi}(\mathbf{z}|\mathbf{x})||p(\mathbf{z}))
\end{equation} 
where the first term can be considered as a 
data reconstruction term, 
and the second a regularization term 
that constrains the learned potential density $q_{\phi}(\mathbf{z}|\mathbf{x})$ 
by a prior $p(\mathbf{z})$ 
via a Kullback-Leibler (KL) divergence measure. 
In practice, both $q_{\phi}(\mathbf{z}|\mathbf{x})$ and $p_{\theta}(\mathbf{x}|\mathbf{z})$ are assumed to be Gaussian.

\begin{algorithm}[t]
  \label{alg:algorithm1}
  \caption{Image Quality Measure (IQM)}
  \textbf{input: } $n$ Images of size $N$x$N$ \;
  \textbf{output: } IQM for frequency domain image quality measure \;
  Initialize \textit{var} as an empty list \;
  \For{each Image $I \in n$}{%
      $F$ = FT ($I$) $\leftarrow$ Fourier Transform \;
      $F_{c}$ = FTS ($F$) $\leftarrow$ shift the zero-frequency component to the center of the spectrum \;
      $AF$ = $|F_{c}|$ \;
      $M$ = $\max AF$  \;
      $T_{H}$ = total number of pixels with value $>$ threshold, where threshold = $M$ / 100 \;
      IQM = $\frac{T_{H}}{N \cdot N}$ \;
      var $\leftarrow$ IQM, append in the list \;
  }
  IQM = $\frac{1}{n}$ $\sum_{i}^{n}$ var$_{i}$
\end{algorithm}

\section{Approach}
\subsection{Motivation I: Frequency content for image quality}
The data log-likelihood term in (\ref{eq:VAEObj}), with Gaussian observation model, is given as:
\begin{equation}
  \label{eq:VAE1stTerm}
 \log{p_{\theta}(\mathbf{x}|\mathbf{z})} = -\frac{1}{2\sigma_{\theta}^2(z)}\norm{\mathbf{x} - \mu_{\theta}(\mathbf{z})}_{2}^{2}  - \frac{1}{2}\log{2\pi \sigma_{\theta}^2(z)}
\end{equation}
For the ease of computation and preventing the VAE cost from collapsing towards -$\infty$, the Gaussian assumption is further relaxed by lower bounding the variance of the observation model \textit{i.e.,} fixing the variance globally to 1 \cite{dai2019diagnosing}. 
Natural images, however, are non-Gaussian and often characterized by a heavy-tailed distribution \cite{weiss2007makes,olshausen1996emergence}. 
We hypothesize that the textural details, contributing towards sharp and realistic-looking image, are model inadequately with the design choice of Gaussian observation space resulting in an over-smoothed image, a limitation often recognized with VAE based generative models \cite{dai2019diagnosing}.

To empirically validate this hypothesis, we attempt to identify whether a correlation exists between the amount of frequency content in natural images with image quality 
measured by the widely-accepted quantitative metric of FID. FID compares the quality of generated images ($\mathbf{x_{g}}$) with the ground-truth images ($\mathbf{x_{r}}$) using the Fr\'echet distance between two multivariate Gaussians:
\begin{equation}
  \label{eq:VAE1stTerm}
  \text{FID} = \lVert \mu_{r} - \mu_{g} \rVert^{2} + \text{Tr}(\Sigma_{r} + \Sigma_{g} - 2(\Sigma_{r}\Sigma_{g})^{\frac{1}{2}})
\end{equation}
where $\mathbf{x}_{r}$ $\sim$ $\mathcal{N}(\mu_{r}, \Sigma_{r})$ and $\mathbf{x}_{g}$ $\sim$ $\mathcal{N}(\mu_{g}, \Sigma_{g})$ are the 2048-dimensional activation of the Inception-v3 pool3 layer for the real and the generated samples respectively. 
A lower value of FID 
implies a higher sample quality.

To measure image sharpness in the frequency domain, we use the  Image Quality Measure (IQM) \cite{de2013image} across the image samples. Unlike FID, a higher IQM value represents a higher image quality by assessing the value of the frequency component in the image obtained via Fourier transform. 
The algorithm for the IQM is presented in Algorithm \ref{alg:algorithm1}.
 
\begin{figure}[tb]
\begin{center}
  \includegraphics[width=0.47\textwidth]{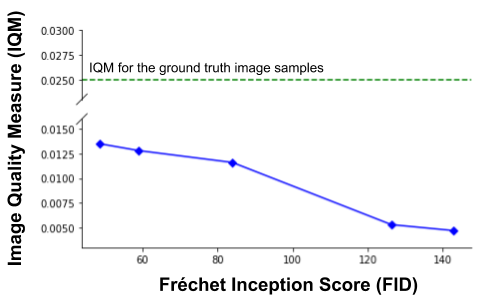}
\end{center}
\vspace{-0.3cm}
\caption{\small{Plot showing correlation between IQM and FID for images generated using different VAE-based generative models trained on CelebA dataset. Note the gap between IQM values of generated samples against ground truth images.}}
\label{fig:IQMvsFID}
\vspace{-0.3cm}
\end{figure}

For multiple VAE-based generative models, we measure both FID and IQM for the CelebA dataset \cite{liu2015deep}. 
As shown in Fig. \ref{fig:IQMvsFID}, a better FID score (lower value) correlates with a higher IQM value.  
This provides important quantitative evidence  that the frequency content of an image plays a crucial role in improving FID score, motivating us to improve the ability of VAE to model the frequency  components  within  an  image.

\subsection{Motivation II: Image priors for decomposing generative operators}
Natural images often lie in a high-dimensional space which, with innovations in image acquisition tools, 
is ever-growing. 
To directly approximate an operator $\hat{F}$ that transforms the latent sample $z \sim q_{\phi}(z|x)$ into the high-dimensional  image $\mathbf{x}$, \textit{i.e.,} $\mathbf{x}$ = $\hat{F}(z)$, 
is difficult and may  
require specialized architectures to capture the associated distributions \cite{karras2019style}. 
Assume that the operator $\hat{F}$ can be decomposed as: 
\begin{equation}
  \label{eq:breakDown}
\hat{F}(z) = (\Phi \odot \hat{G})(z)
\end{equation}
where $\Phi$ is a pre-defined linear operator that maps an intermediate low-dimensional representation of an image to the image itself, and $\hat{G}$ is a ``less'' complex operator.  
With 
a proper decomposition, 
$\hat{G}$ will be simpler to model, 
reducing
the complexity of the generative modeling of $\hat{F}(z)$.
To utilize
prior domain knowledge about the natural images in the decomposition \cite{angles2018generative}, we are motivated to decompose $\hat{F}(z)$ 
into $\Phi$ as a inverse wavelet transform, and $\hat{G}$ as 
a generative model of 
the wavelet coefficients, the latter now to be learned by the wavelet based decoder.

\subsection{Wavelet based encoder-decoder}
We present a probabilistic encoder-decoder network to learn to reconstruct the coefficients of the wavelet decomposition, $\mathbf{y} = \{\mathbf{u}, \mathbf{w}\}$, from the input natural image $\mathbf{x}$. We train this probabilistic model by maximizing the log likelihood 
as:
\begin{equation}
  \label{eq:logliklihood}
\argmax_{\omega}\mathbb{E}_{\mathbb{P(\mathbf{x},\mathbf{y})}} \log p_{\omega}(\mathbf{y}|\mathbf{x})
\end{equation}
where $\mathbb{P(\mathbf{x},\mathbf{y})}$ is the joint distribution of the input image and output wavelet coefficients. We introduce a latent random variable $\mathbf{z}$ and express the conditional distribution as:
\begin{equation}
  \label{eq:conditional}
  p_{\omega}(\mathbf{y}|\mathbf{x}) = \int p_{\theta} (\mathbf{y}|\mathbf{z}) p_{\phi} (\mathbf{z}|\mathbf{x}) d\mathbf{z}
\end{equation}
where $\omega$ = $\{\theta, \phi\}$. In the presented setup we model $p_{\phi} (\mathbf{z}|\mathbf{x})$ with a Gaussian distribution where the parameters of the distribution, mean and variance, are parameterized by deep neural networks:
\begin{equation}
  \label{eq:latentDist}
p_{\phi} (\mathbf{z}|\mathbf{x}) = \mathcal{N}(\mathbf{z}|\mu_{\phi}(\mathbf{x}), \sigma^{2}_{\phi}(\mathbf{x}))
\end{equation}

Considering that wavelet coefficients of natural images are 
found to have more sharply peaked and much heavier tails than a usual Gaussian density \cite{simoncelli1999modeling}, 
we model $p_{\theta} (\mathbf{y}|\mathbf{z})$ with a generalized Laplacian (or "stretched exponential") distribution \cite{simoncelli1996noise}:
\begin{equation}
  \label{eq:generalizedLaplacian}
p_{\theta} (\mathbf{y}|\mathbf{z}) = \frac{e^{-|\mathbf{y}/\mathbf{s}|^{\mathbf{p}}}}{Z(\mathbf{s},\mathbf{p})}
\end{equation}
where $Z(\mathbf{s},\mathbf{p}) = 2\frac{\mathbf{s}}{\mathbf{p}}\Gamma (\frac{1}{\mathbf{p}})$ 
is 
the normalization constant
and $\Gamma (\cdot)$ is the gamma function. 
Here we lower bound the parameters \{$\mathbf{s}, \mathbf{p}$\} of this observation model (\ref{eq:generalizedLaplacian}) by fixing them globally to 1, which reduces the objective to L1-norm between ground-truth observation of wavelet coefficients and the probabilistic approximation of wavelets $\mathbf{y} \sim p_{\theta} (\mathbf{y}|\mathbf{z})$.

With all the distributions involved in (\ref{eq:conditional}) defined, we utilize the variational framework, similar to VAE, for the proposed wavelet encoder-decoder to formulate the objective function into:
\begin{equation}
  \label{eq:WaveObj}
  \mathcal{L} = \mathop{\mathbb{E}_{q_{\phi}(\mathbf{z}|\mathbf{x})}}[\log{p_{\theta}(\mathbf{y}|\mathbf{z})}] - KL(q_{\phi}(\mathbf{z}|\mathbf{x})||p(\mathbf{z}))
\end{equation}
where the first term is the wavelet reconstruction term, and the second is the regularization that constrains the learned potential density from natural images, $q_{\phi}(\mathbf{z}|\mathbf{x})$, by a prior $p(\mathbf{z})$ via Kullback-Leibler (KL) divergence measure. 

The first term of the objective presented in (\ref{eq:WaveObj}) can be decomposed into low-frequency approximation coefficient and high-frequency wavelet coefficients as:
\begin{equation}
  \label{eq:WaveObj1}
\begin{aligned}
 \mathcal{L} = & \mathop{\mathbb{E}_{q_{\phi}(\mathbf{z}|\mathbf{x})}}[\log{p_{\theta}(\mathbf{u}|\mathbf{z})}] +  \beta \cdot \mathop{\mathbb{E}_{q_{\phi}(\mathbf{z}|\mathbf{x})}}[\log{p_{\theta}(\mathbf{w}|\mathbf{z})}] - \\
& KL(q_{\phi}(\mathbf{z}|\mathbf{x})||p(\mathbf{z})) 
\end{aligned}
\end{equation}
where $\beta$ is a tunable hyperparameter contributing towards the importance of the wavelet coefficients. For the low-frequency approximation coefficient $\mathop{\mathbb{E}_{q_{\phi}(\mathbf{z}|\mathbf{x})}}[\log{p_{\theta}(\mathbf{u}|\mathbf{z})}]$, considering its similarity to the image, we make Gaussian assumption and use the pixel-wise mean square loss. For the high-frequency wavelet coefficients $\mathop{\mathbb{E}_{q_{\phi}(\mathbf{z}|\mathbf{x})}}[\log{p_{\theta}(\mathbf{w}|\mathbf{z})}]$, we make Laplacian assumption and use the L1-norm for the wavelet reconstruction.

\begin{figure}[tb]
\begin{center}
  \includegraphics[width=0.47\textwidth, height=0.25\textheight]{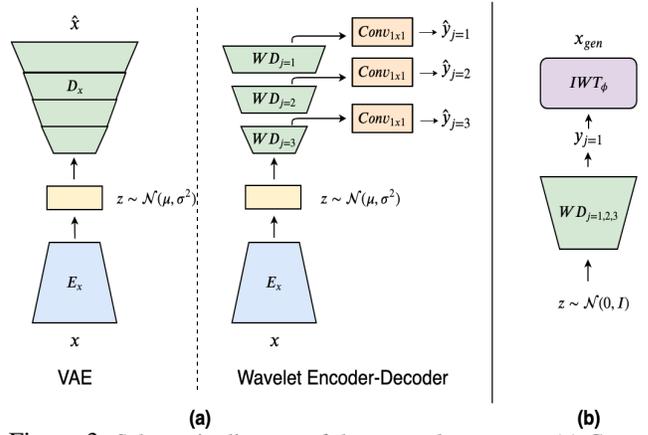}
  \vspace{-1.0cm}
\end{center}
\caption{\small{Schematic diagram of the network structure. (a) Comparison between vanilla-VAE and Wavelet encoder-decoder. The encoder $E_{x}$ are identical for both the models. VAE uses the decoder $D_{x}$ to model the sampled latent code directly to image space, whereas in wavelet encoder-decoder, the decoder $WD_{j=i}$ is used to model the latent code to wavelet coefficient space. (b) Image generation pipeline using trained wavelet generative model. The generated wavelet coefficients are passed through Inverse Wavelet Transform ($IWT_{\phi}$) to obtain the RGB image.}} 
\label{fig:network}
\vspace{-0.3cm}
\end{figure}

\subsubsection{Network Structure}
 Fig. \ref{fig:network} (a) presents the schematic diagram of the network structure of the proposed wavelet encoder-decoder (right) in comparison with the standard VAE (left). The encoder $q_{\phi}(\mathbf{z|x})$ in these two models are identical.
The decoders are different since we aim to model scale and wavelet coefficients in the output space $p_{\theta}(\mathbf{y|z})$. 
However, from an architectural perspective, the wavelet decoder is close to the image decoder. 
For example, the same convolution block (denoted by four blocks for VAE and three blocks for wavelet encoder-decoder in Fig. \ref{fig:network} (a)) can be used to build both decoders. 
As the highest level of decomposition $\mathbf{y}_{j=1}$ is scaled down (in half) compared to the image $\mathbf{x}$, in wavelet decoder, the final block of image decoder is not required. At the same time, since the scale and the wavelet coefficients comprise in total four different components: {LL, LH, HL, HH}, the output channel size (compared to image decoder) is increased by a factor of four and thus, we add point-wise convolution (\textit{i.e.}, kernel = 1x1) to meet this requirement.
As an example, for a generative model for an image $\mathbf{x} \in \mathbb{R}^{3\times 64\times 64}$, the first level wavelet decomposition would be $\mathbf{y}_{j=1} \in \mathbb{R}^{12\times 32\times 32}$, the second level wavelet decomposition would be $\mathbf{y}_{j=2} \in \mathbb{R}^{12\times 16\times 16}$, etc.

As has been mentioned before, \textit{multi-resolution} is one of the important attributes of the wavelet decomposition. To utilize this primary property, we propose a network structure capable of handling such hierarchical multi-resolution wavelets. Since different convolution blocks within the decoder output the feature maps at different resolutions, we utilize them to represent the wavelet decomposition at different scales as shown in Fig. \ref{fig:network} and use each of them in the data likelihood term as presented in (\ref{eq:WaveObj1}).
To evaluate the use of such multi-resolution decomposition we experiment with two versions of the network structure: 1) decoding three levels of wavelet decomposition ($\mathbf{y}_{j=1,2,3}$) and 2) decoding a single level of wavelet decomposition ($\mathbf{y}_{j=1}$).

\begin{table*}[t]
  \centering
  \begin{tabular}{|c|c|c|c|c|c|c|}
  \hline
    Model & \multicolumn{3}{c|}{Generated Images} & \multicolumn{3}{c|}{Reconstructed Images}  \\
    & CIFAR-10 & CelebA & Flickr-Face & CIFAR-10 & CelebA & Flickr-Face \\
    \hline
    VAE & 130.62 $\pm$ 0.37 &  57.19 $\pm$ 0.09 & 79.35 $\pm$ 0.32 & 122.54 $\pm$ 0.09 &  48.57 $\pm$ 0.05 & 73.56 $\pm$ 0.24 \\
    
    VAE-c & 104.35 $\pm$ 0.31 & 53.87 $\pm$ 0.16 & 80.52 $\pm$ 0.14 & 92.23 $\pm$ 0.21 & 40.78 $\pm$ 0.03 & 70.96 $\pm$ 0.10 \\
    
    Wavelet-VAE & 112.45 $\pm$ 0.17 &  49.96 $\pm$ 0.17 & \textbf{71.38} $\pm$ 0.14 & 92.40 $\pm$ 0.12 & 43.94 $\pm$ 0.08 & \textbf{65.72} $\pm$ 0.07 \\
    
    Wavelet-VAE-MR & \textbf{101.86} $\pm$ 0.28 &  \textbf{47.92} $\pm$ 0.08 & 84.90 $\pm$ 0.28 & \textbf{80.65} $\pm$ 0.13 & \textbf{37.89} $\pm$ 0.07 & 71.21 $\pm$ 0.14\\
    \hline
  \end{tabular}
  \caption{FID score comparison for \textit{generated} and \textit{reconstructed} images of the presented Wavelet-VAE and Wavelet-VAE-MR models against standard VAE trained with MSE loss and VAE trained with cross-entropy loss for the likelihood term. For each dataset, we performed 5 independent trials and report the mean and standard deviation of the FID scores.}
  \label{tab:genFID}
  \vspace{-0.3cm}
\end{table*}


\begin{figure}[tb]
\begin{center}
  \includegraphics[width=0.47\textwidth]{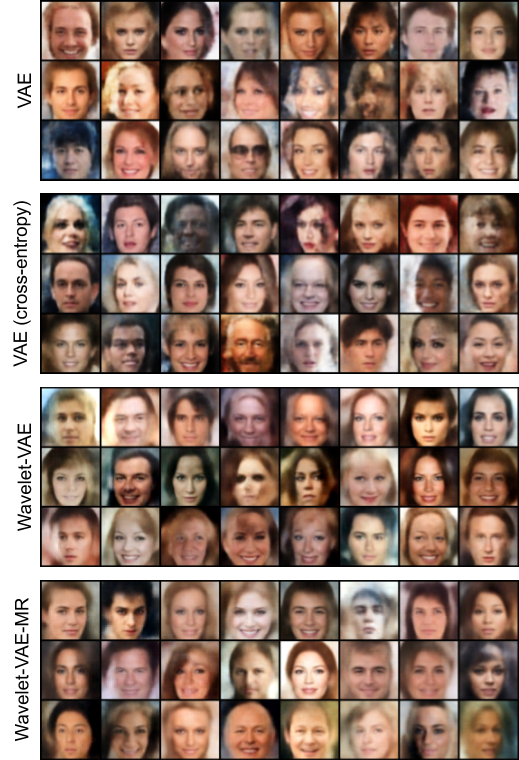}
\end{center}
\vspace{-0.3cm}
\caption{\small{Qualitative comparison of images generated using the proposed Wavelet-VAE and Wavelet-VAE-MR models against standard VAE model and VAE trained using a cross-entropy loss for the likelihood term for CelebA dataset. The images are presented without any cherry-picking.}} 
\label{fig:celebA_basic}
\end{figure}

\begin{figure}[tb]
\begin{center}
  \includegraphics[width=0.47\textwidth]{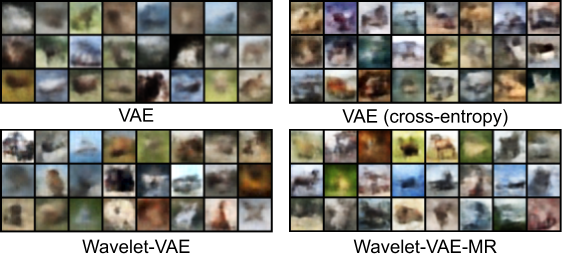}
\end{center}
\vspace{-0.3cm}
\caption{\small{Qualitative comparison of images generated using the proposed Wavelet-VAE and Wavelet-VAE-MR models against standard VAE model and VAE trained using a cross-entropy loss for the likelihood term for CIFAR-10 dataset. The images are presented without any cherry-picking.}} 
\label{fig:cifar}
\vspace{-0.2cm}
\end{figure}

\subsubsection{Image Generation}
Our ultimate goal is to achieve an operator $\hat{F}$ for the generative model $p(\mathbf{x}|\mathbf{z})$. 
With the decomposition proposed in (\ref{eq:breakDown}), 
we now have learned $\hat{G}$ as the decoder, \textit{i.e.,} $p_{\theta}(\mathbf{y}|\mathbf{z})$, in the proposed wavelet-based encoder-decoder.
Since the intermediate representation achieved by $\hat{G}$ is nothing but the space of wavelets, we require $\Phi$ in (\ref{eq:breakDown}) to be an operator that maps the wavelet space back to image space. Inverse wavelet transform combines the approximation coefficients $\mathbf{u}$ and wavelet coefficients $\mathbf{w}$ and inverts them to the image space as shown in (\ref{fig:wavletDecomposition}) and thus essentially is the good candidate for $\Phi$. 

The whole image generation process is schematically shown in Fig. \ref{fig:network} (b) and can be summarized as:

\begin{equation}
  \label{eq:imageGeneration}
\begin{aligned}
& \mathbf{z} \sim \mathcal{N}(0,1) \\
& \mathbf{y} \sim p_{\theta}(\mathbf{y}|\mathbf{z}); \quad \mathbf{y} = \{\mathbf{u}, \mathbf{w}\} \\
& \mathbf{x} = \Phi(\mathbf{u}, \mathbf{w})
\end{aligned}
\end{equation}
The image generation process can also utilize the different levels of $\mathbf{y}$ generated by the wavelet decoder. In this work,
we only considered the image generated via highest level prediction of coefficients \textit{i.e.,} $\mathbf{y}_{j=1}$, regardless of the number of levels used while training the wavelet encoder-decoder. For brevity, throughout the remainder of this paper, we refer to the pipeline of image generation via wavelet-based encoder-decoder, depending on levels of resolution used to train the wavelet encoder-decoder, as Wavelet-VAE (for $\mathbf{y}_{j=1}$) and Wavelet-VAE-MR (multi-resolution for $\mathbf{y}_{j=1,2,3}$).



\section{Experiments}
In this section, we evaluate the generation and reconstruction capabilities of the proposed Wavelet-VAE qualitatively and quantitatively. For quantitative evaluations, we use FID scores \cite{heusel2017gans}. We consider three popular natural image datasets: CIFAR-10 \cite{krizhevsky2009learning}, CelebA \cite{liu2015deep}, and Flickr-Face dataset \cite{karras2019style}. We start with comparing Wavelet-VAE against two baseline image-space VAE models differing only in the data likelihood term of ELBO. The first one is a vanilla-VAE with Gaussian observation space (VAE trained with mean-squared loss) with fixed variance \cite{kingma2013auto}. Recent works \cite{dai2019diagnosing} have demonstrated superior reconstruction and generation quality when cross-entropy is used as the likelihood term. We use this as the second baseline and refer to it as ``VAE-c''. We further analyze the properties of factor disentanglement and relevance of information embedded in the latent space. Finally, we use wavelet-space decoder as the generator in GANs to design wavelet-GAN and compare with image-space GAN for a variety of the GAN models.

\subsubsection{Implementation Details}
In this work, we choose the simple Haar wavelet as the basis function for the 2D discrete wavelet transform (DWT) and the inverse wavelet transform (IWT). 
For a fair comparison, we use standard experimental settings from \cite{dai2019diagnosing} for all the networks and training processes without any arbitrary fine-tuning and adopt a common neural architecture for all models with the encoder and decoder networks based on the InfoGAN \cite{chen2016infogan}. 
We use a batch size of 100 and set the initial learning rate as 0.0001. For CIFAR-10 dataset, we train the network for 1000 epochs and halve the learning rate every 300 epochs and set the $\beta$ as 5. For CelebA and Flickr-Face datasets, we train the models for 120 epochs with halving the learning rate every 48 epochs and set the $\beta$ as 1. We fix the size of the latent dimension as 64 for all our experiments.

\begin{figure}[t]
\begin{center}
  \includegraphics[width=0.47\textwidth]{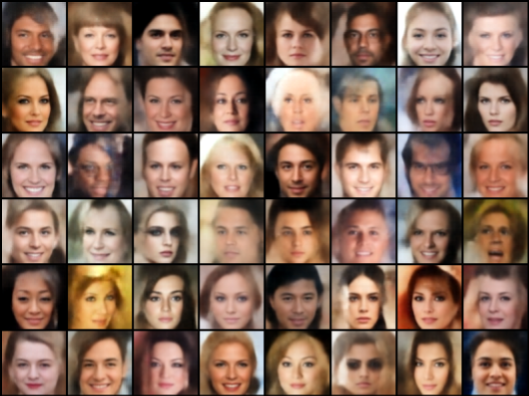}
\end{center}
\vspace{-0.3cm}
\caption{\small{Generated samples from Wavelet-VAE-MR model which uses the architecture and experimental setup from \cite{tolstikhin2017wasserstein}. The FID score obtained for this architecture is \textbf{41.13} which is better than most of the GANs and VAEs as reported in \cite{lucic2018gans,dai2019diagnosing}.}} 
\label{fig:celebA}
\end{figure}

\begin{figure}[tb]
\begin{center}
  \includegraphics[width=0.47\textwidth]{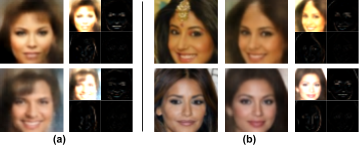}
\end{center}
  \vspace{-0.3cm}
\caption{\small{Visualizing generated and reconstructed scale and wavelet coefficients $\mathbf{y}$. (a) Two generated images along with their associated scale and wavelet coefficients are presented. (b) Two original images, corresponding reconstruction and associated scale and wavelet coefficients are presented. In both (a) and (b), ordering of $\mathbf{y}$ is same with scale coefficient placed at top-left and wavelet coefficients placed in remaining three quadrants.}}
\label{fig:frequency}
\vspace{-0.2cm}
\end{figure}

\begin{figure}[tb]
\begin{center}
  \includegraphics[width=0.50\textwidth]{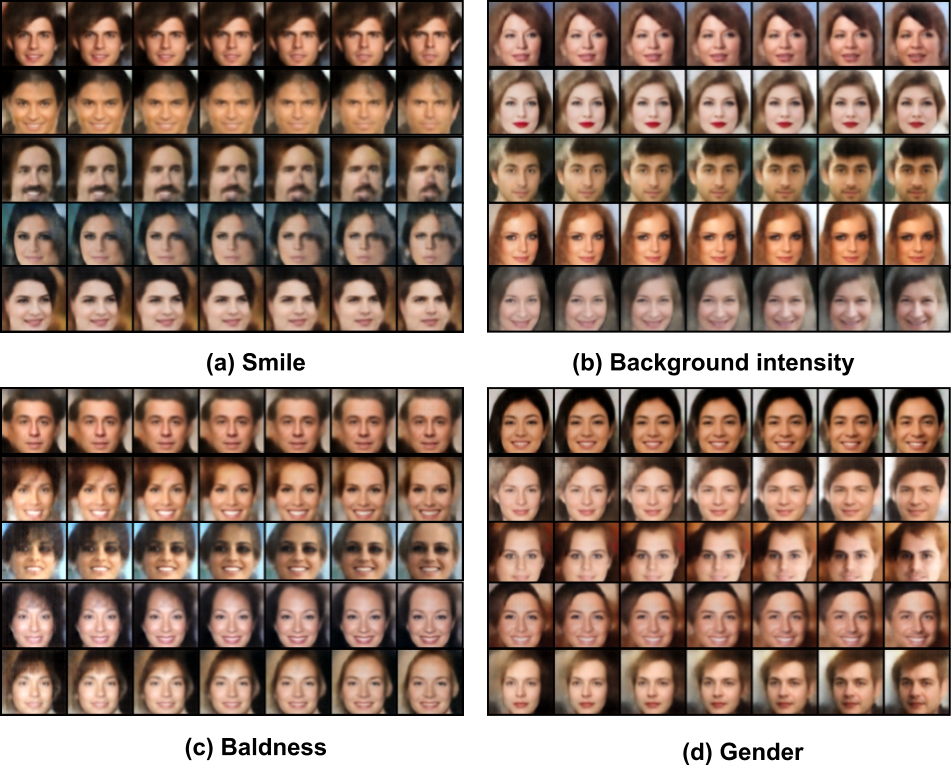}
\end{center}
\vspace{-0.3cm}
\caption{\small{Latent space traversal for Wavelet-VAE-MR model trained for CelebA dataset depicting disentanglement of different generative factors without losing the sharp image details.}} 
\label{fig:disnentgl}
\end{figure}


\subsection{Results}
\subsubsection{Quality of generated natural image samples}
In Table \ref{tab:genFID}, we present the FID score for the presented Wavelet-VAE compared against the two standard VAE models for both generated and reconstructed image samples. For all three natural image datasets, the Wavelet-VAE is consistently better than both the baseline VAE models. 
Among the two out of three datasets, we achieved better performance using the multi-resolution version of the proposed Wavelet-VAE.
For qualitative comparison, we present randomly generated samples from CelebA and CIFAR-10 dataset in Fig. \ref{fig:celebA_basic} and Fig. \ref{fig:cifar} respectively. 
Furthermore, we experimented with a different architectural setting taken from \cite{tolstikhin2017wasserstein} for CelebA dataset and obtained impressive qualitative and quantitative performance. In particular, we obtained an FID score of \textbf{41.13} for generated samples which is better than most of the state-of-the-art VAEs and GANs as reported in \cite{lucic2018gans,dai2019diagnosing}. We present the qualitative results in Fig. \ref{fig:celebA}.

In Fig. \ref{fig:frequency}, we visualize the generated and reconstructed coefficients $\mathbf{y}$ = $\{\mathbf{u}, \mathbf{w}\}$. The scale coefficient $\mathbf{u}$, as expected, captures the scaled approximation of the image and the wavelet coefficients $\mathbf{w}$ capture the different structure and textural information of the image. In particular, the top-right patch, $\mathbf{\psi}_{b=HL}$, captures horizontal frequency (edges) components relating to mouth lining, eye-lids etc. On the other hand, the bottom-left patch, $\mathbf{\psi}_{b=LH}$, captures vertical frequency components relating to facial outline, nose lining, etc. The bottom right patch, $\mathbf{\psi}_{b=HH}$, supposed to capture diagonal frequency components, in our case, however, is not visibly present. 


\begin{table}[t]
  \centering
  \begin{tabular}[t]{|c|c|c|}
    \hline
    Model & Image-GAN  & Wavelet-GAN \\
    & \cite{lucic2018gans} & \\
    \hline
    NS GAN & \textbf{55.0} $\pm$ \textbf{3.3} & 60.30 $\pm$ 1.2  \\ 
    LS GAN & \textbf{53.9} $\pm$ \textbf{2.8} & 54.5 $\pm$ 4.0  \\ 
    WGAN & 41.3 $\pm$ 2.0 & \textbf{40.2} $\pm$ \textbf{1.4}  \\ 
    WGAN GP &  30.3 $\pm$ 1.0 & \textbf{28.7} $\pm$ \textbf{1.2}  \\ 
    \hline 
  \end{tabular} 
  \caption{\small{FID comparision for generated images of the wavelet-spcae GAN against the standard image-space GAN for CelebA dataset. The reported values for image-space GAN represent the optimal FID obtained across a large-scale
hyperparameter search with outliers (e.g., severe mode collapse) removed before calculation \cite{lucic2018gans}. The values for wavelet-space GAN are obtained using a DCGAN based architecture without removing any outliers. Specialized architectures and/or training protocol can potentially improve the results.}}
  \vspace{-0.3cm}
  \label{tab:GANFID}
\end{table}

\subsubsection{Disentanglement and informative latent space}
The tension between good disentanglement and sound reconstructions/generation is a well-observed phenomenon across different state-of-the-art disentanglement methods \cite{higgins2017beta,chen2018isolating}. For the CelebA dataset, as qualitatively shown in Fig. \ref{fig:disnentgl}, the disentanglement achieved by Wavelet-VAE does not seem to lose the sample quality compared to the qualitative results presented in prior works. 

The change in the decoder function has often yielded in problems of posterior collapse or uninformative latent space in VAE \cite{he2019lagging}. To inspect the use of latent space, we calculate the mutual information (MI) $I_{q}(\mathbf{z}; \mathbf{x})$ between the data variable and the latent variable. We utilize the index-code mutual information from ELBO decomposition as proposed in \cite{chen2018isolating} and calculate it using all the training dataset (162771 for CelebA dataset). For the Wavelet-VAE-MR, we obtained a mutual information of 12.10 nats compared to 11.98 nats for the Wavelet-VAE. The standard VAE models, VAE and VAE-c, respectively obtained 11.99 nats and 11.98 nats. Besides the mutual information for Wavelet-VAE being slightly better or similar, this result, most importantly, validates that the wavelet encoder-decoder is indeed learning a latent variable model with informative latent space, an attribute often enjoyed by VAE based models.


\subsubsection{Comparison with GAN-based model}
Since the decoder of wavelet-based VAE is architecturally similar to the generator of GAN, we also investigate the possibility of using a wavelet space generator. For CelebA dataset, we compare the wavelet-based GAN with the standard image space generator and discriminator. We consider numerous popular GAN models, including NS GAN \cite{fedus2017many}, LS GAN \cite{mao2017least}, WGAN \cite{arjovsky2017wasserstein} and WGAN-GP \cite{gulrajani2017improved} based on common DCGAN architecture \cite{radford2015unsupervised}. 
For a fair comparison, we followed the training protocols from \cite{lucic2018gans} for the wavelet-GAN.
In Table \ref{tab:GANFID}, we present the FID score obtained from different GAN-models for wavelet-space GAN along with the corresponding image-space GAN. As we can see, the wavelet-space GAN models are competitive with standard image-space GAN models with the best results obtained for the WGAN-GP setting.

\textbf{Limitations and Future work}: 
Our work in this paper has some limitations. We have used network setups and hyperparameters similar to existing image-space VAEs; adapted and trained them to generate wavelet coefficients without any hyperparameter tuning. As seen in Fig. \ref{fig:celebA}, better architectures reduced the FID presented in Table \ref{tab:genFID} by around 7 points.  This motivates us to further explore the best architectures and hyperparameters for the wavelet-space generative models. Also, hierarchical decompostion process of wavelet transforms naturally motivates the need for hierarchical distributions in the latent space which is lacking in our current set of experiments. Moreover, modeling in wavelet-space gives rise to a multi-stage loss objective in ELBO that passes gradients to all the decoding layers simultaneously. This should improve the stability and convergence of GAN training processes. 
Although we have observed better stable training curves with wavelet-space GANs, rigorous experiments are needed to make better claims.   We explore these for the future. 

\section{Conclusion}

In this work, we introduced a novel Wavelet-VAE that generates high-quality image samples. Specifically, the decoder learns to generate wavelet coefficients with increasing scales of frequencies which can then be transformed to image-space using the classical inverse wavelet transform.  Through extensive experiments on multiple benchmarks, we showed superior generation and reconstruction capabilities of the proposed model over existing models. 
We also empirically demonstrated that wavelet-space decoder helps to learn richer latent spaces in terms of disentanglement properties and mutual information with observation space (data). 
Wavelet decoder, when used as the generator network in several GAN architectures achieved competitive or better performance compared to the baseline GANs for the FID metric. With the consistently better performance of the proposed wavelet paradigm across multiple architectures and datasets, we view our framework as a starting point for a new research direction in the field of generative models for images that explicitly emphasizes on high-frequency details of the natural images. 

\bibliographystyle{aaai}
\bibliography{refs}

\end{document}